\documentclass[sigconf]{acmart}

\copyrightyear{2020}
\acmYear{2020}
\setcopyright{acmlicensed}
\acmConference[MM '20]{Proceedings of the 28th ACM International Conference on Multimedia}{October 12--16, 2020}{Seattle, WA, USA}
\acmBooktitle{Proceedings of the 28th ACM International Conference on Multimedia (MM '20), October 12--16, 2020, Seattle, WA, USA}
\acmPrice{15.00}
\acmDOI{10.1145/3394171.3413643}
\acmISBN{978-1-4503-7988-5/20/10}

\graphicspath{{Figure/}{pictures/}{images/}{./}}

\begin{document}
\fancyhead{}

\title{Modeling Caricature Expressions by 3D \\ Blendshape and Dynamic Texture}

\author{Keyu Chen}
\affiliation{\institution{University of Science and Technology of China}}
\email{cky95@mail.ustc.edu.cn}

\author{Jianmin Zheng}
\affiliation{\institution{Nanyang Technological University}}
\email{ASJMZheng@ntu.edu.sg}

\author{Jianfei Cai}
\affiliation{\institution{Monash University}}
\email{Jianfei.Cai@monash.edu}

\author{Juyong Zhang}
\affiliation{\institution{University of Science and Technology of China}}
\email{juyong@ustc.edu.cn}

\begin{abstract}
 The problem of deforming an artist-drawn caricature according to a given normal face expression is of interest in applications such as social media, animation and entertainment. This paper presents a solution to the problem, with an emphasis on enhancing the ability to create desired expressions and meanwhile preserve the identity exaggeration style of the caricature, which imposes challenges due to the complicated nature of caricatures. The key of our solution is a novel method to model caricature expression, which extends traditional 3DMM representation to caricature domain. The method consists of shape modelling and texture generation for caricatures.  Geometric optimization is developed to create identity-preserving blendshapes for reconstructing accurate and stable geometric shape, and a conditional generative adversarial network (cGAN) is designed for generating dynamic textures under target expressions. The combination of both shape and texture components makes the non-trivial expressions of a caricature be effectively defined by the extension of the popular 3DMM representation and a caricature can thus be flexibly deformed into arbitrary expressions with good results visually in both shape and color spaces. The experiments demonstrate the effectiveness of the proposed method.
\end{abstract}

\ccsdesc[500]{Applied computing~Media arts}
\ccsdesc[500]{Computing methodologies~Image manipulation}

\keywords{Caricature; Expression; Deformation; Blendshape; Texture}

\maketitle

\section{Introduction}

Caricatures are non-photorealistic images that exaggerate or simplify the features of subjects, serving as a descriptive art form in a wide range of applications such as social network, animation and entertainment industry. Like most other art forms, caricatures are commonly drawn by artists with subjectivity. Even for the same subject, different artists may focus on different features and thus create caricature portraits different in image styles and exaggeration forms. Such characteristics ensure the fascination of caricatures and meanwhile impose challenges in re-creation and editing such as manipulating the expressions of artist-created caricature paintings.  

This paper considers the problem of deforming a 2D caricature image driven by an expression given in a normal face image. The deformed or re-created caricature is expected to exhibit the desired expression and also preserve the identity exaggeration style of the original caricature. In order to create caricatures by computer, many works have been developed to mimic artists’ creative drawing process. They focus on automatic caricature generation in 2D~\cite{Shi2018WarpGANAC,Li2018CariGANCG,Han2018CaricatureShopPA,Cao2018CariGANsUP} and 3D~\cite{Wu2018AliveCF,OToole1997ThreedimensionalCO,Clarke2011AutomaticGO} as well. Other related processes for caricatures such as art style classification~\cite{Yaniv2019TheFO}, caricature landmark detection~\cite{Yaniv2019TheFO,Stricker2018FacialLD} and caricature identity recognition~\cite{Huo2018WebCaricatureAB} have also been developed. With the advance of machine learning techniques, generative adversarial networks (GANs) are designed to relate normal human face domain and caricature domain in caricature generation~\cite{Cao2018CariGANsUP, Shi2018WarpGANAC, Li2018CariGANCG}. For example, the ``face-to-caricature’’ generation builds an unpaired mapping function between two different sets. Different from these works, our work is a problem of deformation, a process on an existing caricature, rather than a creation from scratch. Moreover, our deformation is driven by a facial expression, which is a kind of semantic-guided manipulation natural to human's creation activity. To best of our knowledge, there is not much work done along this direction in the caricature domain. 

Existing caricature techniques do not solve our problem well. Our problem has three technical challenges. First, caricatures generally do not have explicit expression definition or representation. Therefore when we extend the human facial expression representation~\cite{Ekman1978FacialAC}  to the target caricature domain, we have to carefully and properly design the map. Second, the expression deformation often loses the preservation of the caricature’s exaggeration styles. Third, large expression deformation often introduces artifacts in local regions such as the mouth and eyes’ area.

Considering that the expression varies in both geometry and texture spaces, we propose a solution to overcome the challenges by constructing a series of 3D caricature expression blendshapes for shape deformations and a conditional GAN for dynamic texture generation. Specifically, we map the source caricature landmarks to the human face domain using \textit{CariGeoGAN}~\cite{Cao2018CariGANsUP} and obtain a group of blendshapes via fitting a 3DMM model~\cite{Blanz1999AMM,Zhu2016FaceAA}. Then we transfer the landmarks on the normal faces back to the caricature domain and reconstruct 3D dense meshes. In this way, the caricature expressions can be defined by the expression coefficients of 3DMM~\cite{Blanz1999AMM}.  To maintain the identity consistency and the exaggeration style, we propose a geometric optimization procedure to reconstruct the caricature blendshapes from the mapped landmarks by constraining caricature expressions to keep the same exaggeration style.  To eliminate artifacts generated in the deformation, we train a texture generation network conditioned on the expression coefficients. Inspired by~\cite{Pumarola2018GANimationAF}, we adopt the attention module to learn which area should be modified in color space. In order to improve the robustness in dealing with different identities and color styles, we transfer many human face textures to caricature styles by adaptive instance normalization (AdaIN)~\cite{Huang2017ArbitraryST}. Our network is trained in a supervised way on the adapted texture dataset and trained in an unsupervised way on a caricature texture dataset. The combined training manner helps to overcome the lack of a large caricature dataset. In the inference stage, the network takes as input the source texture with target expression parameters and outputs the desired texture for the target expression. As a result, we come up with a novel caricature expression generation method that consists of a shape component and a texture component.

The contribution of this paper lies in three aspects. First, we propose a framework consisting of shape modeling and  texture generation to tackle the challenges occurred in expression driven caricature deformation. 
Second, we design an optimization-based method to construct and refine 3D caricature blendshapes, which extends the popular facial expression representation to the caricature domain. 
Third, we present a dynamic texture generation module for caricatures by training a GAN model conditioned on expression parameters.

\vspace{-0.1in}
\section{Related Work}
This section briefly reviews some related works: 
caricature generation, blendshape representation and facial expression manipulation.

\vspace{-0.2in}
\subsection{Caricature Generation}
Caricature generation aims at creating exaggerated caricatures from given human portraits. Techniques in this area can be classified into two categories: 2D image-based and 3D geometry-based approaches. 2D image-based caricature generation was firstly introduced with interactive methods~\cite{Brennan1985CaricatureGT,Sadimon2010ComputerGC}. With advances in image processing, some automatic approaches were developed. For example, \citet{liao2004automatic} developed a caricature generation system that automatically analyzes the facial features of a subject and then determines how the face components should be altered and placed. Recently, deep learning is adopted into the generation process. \citet{Cao2018CariGANsUP} proposed a shape and texture exaggeration process by using CycleGAN structure. \citet{Shi2018WarpGANAC} developed an automatic network for generating caricatures, which learns how to warp a face photo into a caricature and transfer texture styles as well. Usually the shape and texture processes used in these methods to control the generation are independent.

Relatively, the work for 3D geometry-based caricature generation is much less. Despite some surface based methods~\cite{o19993d,OToole1997ThreedimensionalCO,Sela2015ComputationalCO,mustafa2015L1}, only a few works associate the 3D shape generation process (i.e., caricature reconstruction) with 2D inputs. \citet{Wu2018AliveCF} introduced an intrinsic deformation representation for constructing caricatures from 2D images. \citet{Han2018CaricatureShopPA} proposed a method for generating personalized caricatures from 2D sketches. Compared to generating 2D caricatures, generating 3D caricatures requires more control of shapes and textures and thus is more difficult. On the other hand, 3D models provide more information that can benefit other processes such as animation or face editing. 
\vspace{-0.1in}
\subsection{Blendshape Representation}
Blendshape is a prevalent parametric model in animation and movie industry, which represents facial  variations as linear combinations of several given shapes~\cite{Guenter1998MF}. The basic theory can be dated back to the facial action coding system (FACS)~\cite{Ekman1978FacialAC}. Recently developed 3D face reconstruction works take pre-defined blendshapes (e.g., \ FaceWarehouse model~\cite{cao2014facewarehouse} or 3D morphable model-3DMM~\cite{Blanz1999AMM}) for expression and identity modelling~\cite{deng2019accurate,Guo2017CNNBasedRD,Tran2016RegressingRA,Jiang20183DFR,Gecer2019GANFITGA}. It is also used in various applications such as facial retargeting~\cite{Lewis2014PracticeAT,weise2011realtime} and manipulation~\cite{Orvalho2017TransManipulation}. A blendshape model can be considered as the bases of the expression deformation space of a particular subject. There are several approaches to creating blendshapes. For a real actor, a template model can be registered to its scan and repeating the process for different expressions can generate a range of topology-consistent blendshapes~\cite{cao2014facewarehouse}. For digital characters, a skilled modeling artist can deform a base mesh into different shapes to cover the expression range~\cite{Lewis2014PracticeAT}. 

\vspace{-0.1in}
\subsection{Facial Expression Manipulation}
Many facial expression manipulation methods are based on 3D parametric facial models such as 3DMM. For example, \citet{Vlasic2005FaceTW} proposed a multilinear model for tracking and retargeting expression information. \citet{Cao2014DisplacedDE} extended this idea by performing co-tensor computation on FaceWarehouse dataset~\cite{cao2014facewarehouse}. \citet{Thies2018Face2FaceRF} proposed video-to-video facial expression retargeting by fitting 3DMM with additional lighting parameters.

However, 3D parametric representation is not capable of representing fine-scale geometry and texture details such as teeth and wrinkles. Therefore, many recent approaches editing 2D facial images are based on generative adversarial networks (GANs)~\cite{Goodfellow2014GenerativeAN}. For instance, by considering the expression generation process as unpaired image-to-image translation~\cite{Zhu2017UnpairedIT}, \citet{Pumarola2018GANimationAF} designed an unsupervised framework \textit{GANimation} within the CycleGAN structure to train the face generation network conditioned on Face Action Units. \citet{Ververas2019SliderGANSE} replaced the action unit vector with 3DMM coefficients for stable and accurate expression manipulations. \citet{Geng20193DGF} combined the advantages of both geometry-based methods and image generation networks and presented a 3D guided texture and shape refinement approach.

\section{Methodology}\label{sec:Method}
\subsection{Overview}\label{sec:overview}
Our basic problem can be described as follows: given a caricature image $\mathbf{I}(\mathbf{e_{x}})$ with caricature expression $\mathbf{e_x}$ and labelled facial landmarks $\mathbf{L}(\mathbf{e_{x}})$ and any normal face image $\mathbf{\tilde{I}}(\mathbf{e_{t}})$ with expression $\mathbf{e_t}$, our target is to change the caricature expression to $\mathbf{e_t}$, i.e. generating $\mathbf{I}(\mathbf{e_{t}})$. Following the common approach of 3D parametric modelling of normal faces, we aim to build up a parametric caricature expression model that shares the same identity with $\mathbf{I}(\mathbf{e_{x}})$ while covering different expression variations.
With the parametric expression model, any caricature expression can be parametrically represented by the pre-defined bases. However, unlike normal faces, there is no explicit definition on the caricature expression and there is also no sufficient caricature expression data to build up the parametric model bases. Our idea is to associate the caricature expression with the popular normal face expression model \textit{3D Morphable Model} (3DMM)~\cite{Blanz1999AMM} by adopting the cross-domain mapping \textit{CariGeoGAN} from~\cite{Cao2018CariGANsUP}, which facilitates the mapping from a caricature face's landmarks to the corresponding normal face's landmarks that can be modelled by 3DMM or inversely. In this way, we can define all the expressions, e.g. $\mathbf{e_x}$ or $\mathbf{e_t}$, as the 3DMM expression coefficients, which is widely used in human face related applications. 

In particular, we decompose the caricature new expression generation task into a shape modelling component and a texture generation component. Such disentanglement separates the expression deformation in geometry and color space and allows specific optimization for each component. To obtain the initial shape and texture from original caricature data $\{\mathbf{I}(\mathbf{e_{x}}), \mathbf{L}(\mathbf{e_{x}})\}$, we employ the 3D caricature reconstruction method in~\cite{Wu2018AliveCF} to reconstruct the caricature shape $\mathbf{S}(\mathbf{e_{x}})$ from $\mathbf{L}(\mathbf{e_{x}})$. Then we apply classical rasterization and rendering pipeline in graphics to extract the texture map $\mathbf{T}(\mathbf{e_{x}})$ from $\mathbf{I}(\mathbf{e_{x}})$, based on a pre-defined ARAP parameterization map~\cite{Liu2008ALA}. Note that the texture maps are essentially 2D meshes with fixed vertex locations but varying colors corresponding to different textures.

\begin{figure}
\begin{center}
  \includegraphics[width=1\linewidth]{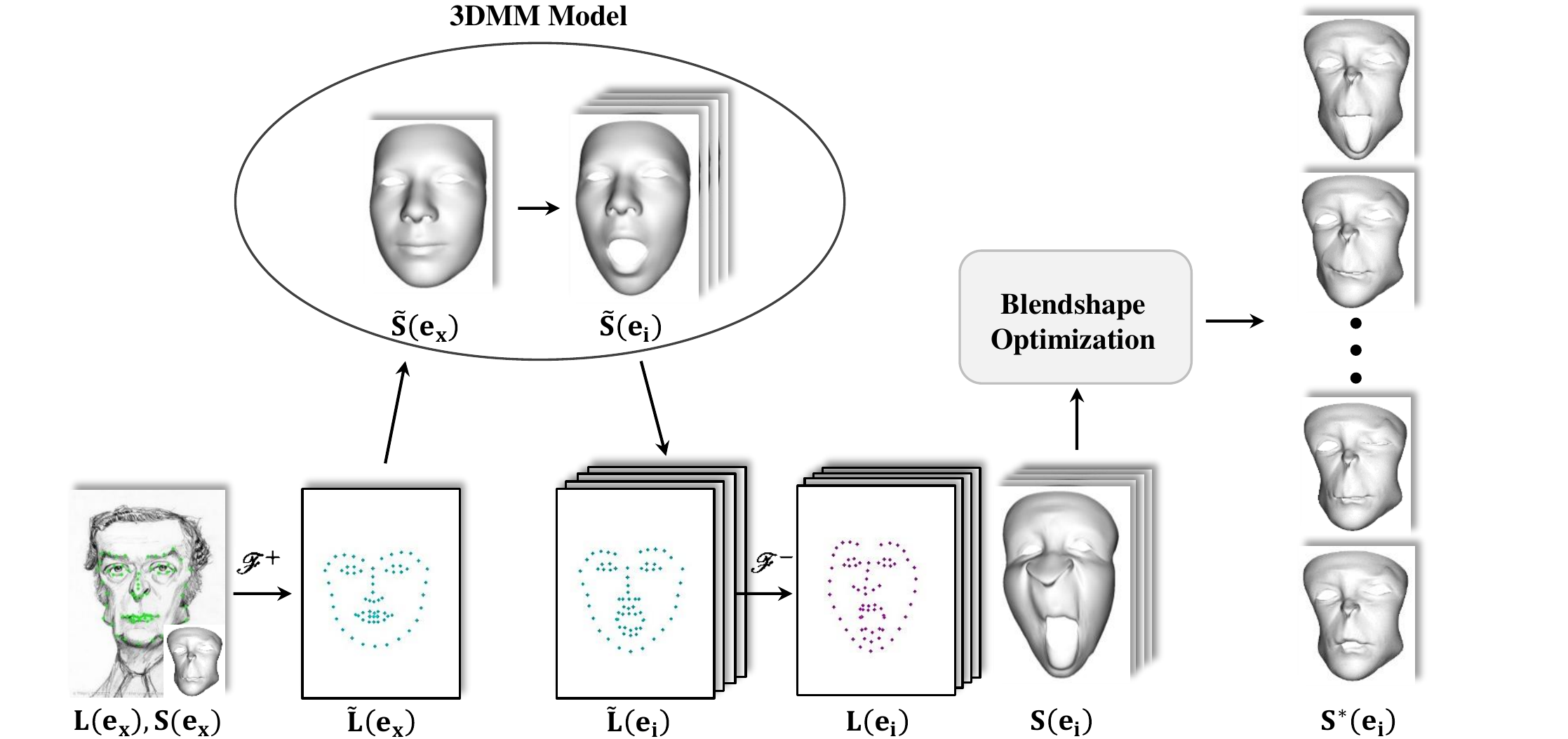}
\end{center}
  \vspace{-0.1in}
  \caption{Process of constructing 3D caricature blendshapes. From left to right, we first fit 3DMM model on the mapped landmarks $\mathbf{\tilde{L}}(\mathbf{e_{x}})$ and then translate those manipulated landmarks $\mathbf{\tilde{L}}(\mathbf{e_{i}})$ back to the caricature domain, obtaining $\mathbf{{L}}(\mathbf{e_{i}})$. The initial caricature blendshapes $\mathbf{{S}}(\mathbf{e_{i}})$ are reconstructed based on~\cite{Wu2018AliveCF} and further optimized by our optimization module for preserving the identity.}
\label{fig:blendshape_construction}
\end{figure}

For the shape modelling component, our target is to construct 3D caricature blendshapes $\{\mathbf{S}(\mathbf{e_{i}})\}_{i=0}^N$ where $\{\mathbf{e_{i}}\}_{i=0}^N$ are expression parameters defined in 3DMM. To this end, we use the cycle consistent network \textit{CariGeoGAN}~\cite{Cao2018CariGANsUP} to translate caricature landmarks $\mathbf{L}(\mathbf{e_{x}})$ to normal face domain and fit a 3DMM model on the mapped landmarks. By changing the expression parameters to $\{\mathbf{e_{i}}\}_{i=0}^N$that are pre-defined in normal face domain, we get the blendshape landmarks, which are then mapped back to the caricature domain for reconstructing 3D caricature blendshapes. During this process, the caricature identity information is often disturbed by expression deformation. Thus, we propose an optimization approach to refine the blendshapes to preserve the identity of original shape $\mathbf{S}(\mathbf{e_{x}})$. The details are illustrated in Fig.~\ref{fig:blendshape_construction} and described in Sec.~\ref{sec:shape_component}.

\begin{figure*}
\begin{center}
  \includegraphics[width=1\linewidth]{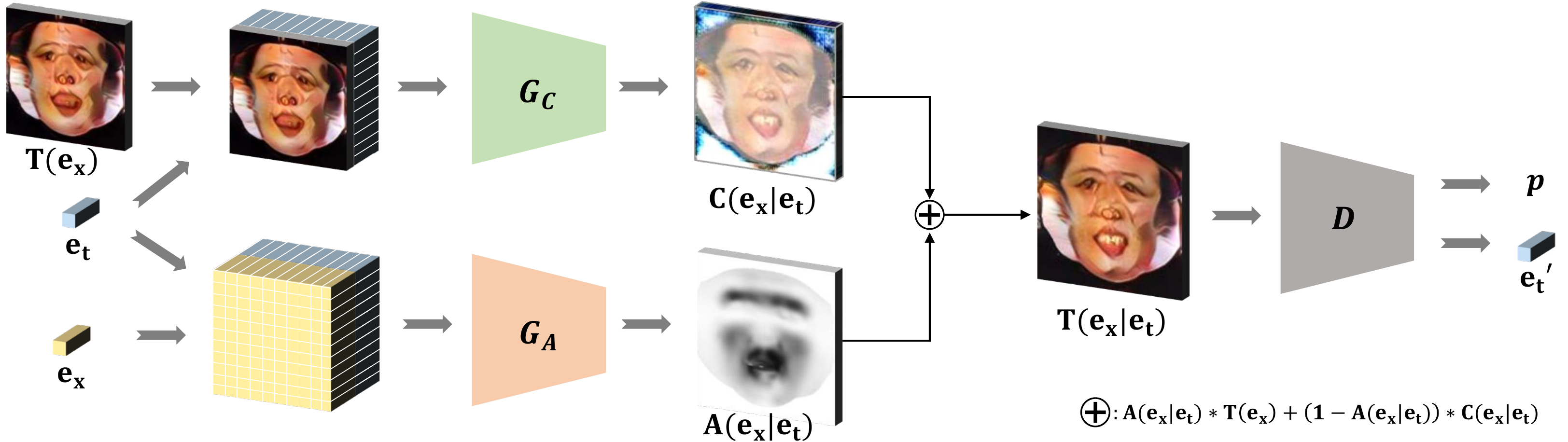}
\end{center}
\vspace{-0.1in}
  \caption{\textbf{The conditional GAN based dynamic texture generation network.} The inputs include the original caricature texture map $\mathbf{{T}}(\mathbf{e_{x}})$, the original expression $\mathbf{e_{x}}$ and the target expression $\mathbf{e_{t}}$, and the output is the generated new texture map $\mathbf{{T}}(\mathbf{e_{x}}|\mathbf{e_{t}})$. The two-branch generator estimates color and attention masks separately, where the former suggests what color values should be changed to and the latter indicates where to make changes. The discriminator is to judge whether the generated texture map looks real or not as well as predicting its expression $\mathbf{e_{t}'}$, which is to be close to $\mathbf{e_{t}}$.} 
\label{fig:network_structure}
\end{figure*}

Meanwhile, just modeling 3D geometric deformation is not enough for image-based caricature expression manipulation. Not only the shapes but also the textures should be deformed in expression changes~\cite{Geng20193DGF}. For example, when a neutral face image is manipulated to be a smile, the teeth part should be added on the mouth area for better photorealistic. Based on this observation, we propose a dynamic texture module for generating different caricature textures under different expression conditions. Our texture generation model is a conditional GAN and importantly it still takes the same 3DMM expression representation as the shape modelling component. 
In the training stage, we overcome the issue of lacking large scale caricature dataset by adapting normal face textures to caricature styles using \textit{Adaptive Instance Normalization}~\cite{Huang2017ArbitraryST} (AdaIN). In the inference stage, our texture model takes as input original texture $\mathbf{T}(\mathbf{e_{x}})$ and target expression $\mathbf{e_{t}}$, and outputs the desired texture map $\mathbf{T}(\mathbf{e_{t}})$. The texture generation can be formulated as $\{\mathbf{T}(\mathbf{e_{x}})|\mathbf{e_{t}}\}\to \mathbf{T}(\mathbf{e_{t}})$. The details are illustrated in Fig.~\ref{fig:network_structure} and described in Sec.~\ref{sec:texture_component}.

\vspace{-0.1in}
\subsection{Shape Modelling Component}\label{sec:shape_component}
In order to faithfully construct 3D blendshapes for the original caricature $\{\mathbf{I}(\mathbf{e_{x}}), \mathbf{L}(\mathbf{e_{x}})\}$, we combine two state-of-the-art methods: 3D caricature reconstruction~\cite{Wu2018AliveCF} and photo-caricature landmark mapping~\cite{Cao2018CariGANsUP} (See Sec.~\ref{sec:preliminary}), to construct initial caricature blendshapes, which are then further refined by our carefully designed optimization module to generate identity-preserving 3D caricature blendshapes.
\subsubsection{Preliminary}\label{sec:preliminary}
\noindent \textbf{3D Caricature Reconstruction} refers to the process of generating 3D caricature model from 2D information. Different from previous deformation-based techniques requiring existing 3D caricature template, Wu~\cite{Wu2018AliveCF} proposed an optimization approach that can reconstruct 3D caricature from sparse 2D landmarks by using intrinsic deformation representation~\cite{Gao2019SparseDD}. Assuming the original caricature landmarks $\mathbf{L}(\mathbf{e_{x}})\in \mathbb{R}^{2\times K}$ labelled on image plane, the method~\cite{Wu2018AliveCF} can output the caricature 3D mesh $\mathbf{S}(\mathbf{e_{x}})\in \mathbb{R}^{3\times N}$ with associated camera parameters $\mathbf{\Pi, R, t}$. The orthographic relationship between 2D and 3D can be written as:
\begin{equation}\label{eq:projection}
\mathbf{q} = \mathbf{\Pi Rp}+\mathbf{t},
\end{equation}
where $\mathbf{\Pi}\in \mathbb{R}^{2\times 3}$ is the scaling matrix, $\mathbf{R}\in \mathbb{R}^{3\times 3}$ is the rotation matrix, $\mathbf{t}\in \mathbb{R}^{2\times 1}$ is the translation vector in image plane, $\mathbf{p}$ and $\mathbf{q}$ are the locations of a 3D dense mesh vertex in the world coordinate system and the image plane, respectively. $K$ and $N$ are vertex numbers of 2D sparse landmarks and 3D dense mesh.

\noindent \textbf{Photo-Caricature Landmark Mapping} serves as cycle translation functions between normal face and caricature domain. In order to automatically generate caricature images from human portraits, Cao~\cite{Cao2018CariGANsUP} proposed a cycleGAN based method to translate facial landmarks across two domains. In our problem, we find that such a cross mapping function bridging caricatures and normal faces can be used to extend the human facial expression representation like 3DMM. In particular, we firstly transfer the original caricature landmarks $\mathbf{L}(\mathbf{e_{x}})$ to normal face domain as $\tilde{\mathbf{L}}(\mathbf{e_x})=\mathcal{F}^{+}(\mathbf{L}(\mathbf{e_x}))$. Then by fitting a 3DMM model on $\tilde{\mathbf{L}}(\mathbf{e_x})$ and manipulating expression parameters to the pre-defined expression as $\mathbf{e_i}$, we project the new 3D landmark points to image plane as $\tilde{\mathbf{L}}(\mathbf{e_i})$. Finally, we map the generated normal facial landmarks back to caricature domain as $\mathbf{L}(\mathbf{e_i})=\mathcal{F}^{-}(\tilde{\mathbf{L}}(\mathbf{e_i}))$. $\mathcal{F}^{+}$ and $\mathcal{F}^{-}$ are the bi-directional translation networks adopted from~\cite{Cao2018CariGANsUP}. Through this way, we can modify the underlying expression of the original caricature landmarks.

\subsubsection{Blendshape Construction}\label{sec:blendshape_construction}
Based on the mapped landmarks in normal face demain, i.e. $\tilde{\mathbf{L}}(\mathbf{e_x})=\mathcal{F}^{+}(\mathbf{L}(\mathbf{e_x}))$, we firstly approximate the PCA-based expression and identity coefficients within 3DMM to fit the 3D face model as:
\begin{equation}\label{eq:3dmm}
    \tilde{\mathbf{S}}(\mathbf{e_{x}}) = \tilde{\mathbf{S}}_{mean} + \mathbf{U}_{id}\cdot \mathbf{\alpha}_{id} + \mathbf{U}_{exp}\cdot \mathbf{\alpha}_{exp},
\end{equation}
where $\tilde{\mathbf{S}}_{mean}$ is the mean face shape, and $\mathbf{U}_{id}$ and $\mathbf{U}_{exp}$ are the PCA bases for identity and expression, respectively. Accordingly, the expression of the source caricature can be directly defined as the expression parameters in Eq.~\ref{eq:3dmm}, i.e., $\mathbf{e_{x}} = \mathbf{\alpha}_{exp}$.

Next we select a group of existing blendshapes from FaceWarehouse~\cite{cao2014facewarehouse} dataset and extract their expression parameters as $\{\mathbf{e_{i}}\}_{i=0}^{46}$. By replacing the expression parameters $\mathbf{\alpha}_{exp}$ with $\mathbf{e_{i}}$, we deform $\tilde{\mathbf{S}}(\mathbf{e_{x}})$ to
\begin{equation}
\begin{split}
    \tilde{\mathbf{S}}(\mathbf{e_{i}}) &= \tilde{\mathbf{S}}_{mean} + \mathbf{U}_{id}\cdot \mathbf{\alpha}_{id} + \mathbf{U}_{exp}\cdot \mathbf{e_{i}},
\end{split}
\end{equation}
and project the landmarks back to image plane as $\{\tilde{\mathbf{L}}(\mathbf{e_{i}})\}_{i=0}^{46}$. 

Finally, we use the inverse mapping function $\mathcal{F}^-$ to transfer the manipulated landmarks $\tilde{\mathbf{L}}(\mathbf{e_{i}})$ to caricature domain as $\mathcal{F}^{-}(\tilde{\mathbf{L}}(\mathbf{e_{i}}))=\mathbf{L}(\mathbf{e_{i}})$. Following the caricature reconstruction method~\cite{Wu2018AliveCF}, we obtain a group of 3D caricature models $\{\mathbf{S}(\mathbf{e_{i}})\}_{i=0}^{46}$ from landmarks $\{\mathbf{L}(\mathbf{e_{i}})\}_{i=0}^{46}$ that convey the faithful expression $\{\mathbf{e_i}\}_{i=0}^{46}$.

The full process of initializing the caricature blendshape model is depicted in Fig.~\ref{fig:blendshape_construction}. It can be observed that despite the expression semantics of the generated caricature model $\mathbf{S}(\mathbf{e_{i}})$, the identity consistency, however, might be violated significantly. The reasons could be: (a) In the photo-caricature landmark translation process, there is no guarantee that the landmarks of different expressions of the same person shall be mapped consistently along with the identity information; (b) Reconstructing 3D caricature model from sparse landmarks is an ill-posed problem, where the 3D identical information could be easily affected by expression deformation in a low-dimensional space. To address this issue, next we propose an optimization module to refine the initial caricature models.

\subsubsection{Blendshape Optimization}\label{sec:blendshape_optimization}
In order to refine the initial caricature blendshapes, we design our optimization energies with two objectives: (i) reducing over-exaggeration; (ii) preserving the structural correlations among blendshape sets.

For the first objective, we adopt the \textit{handle-based surface editing}~\cite{SorkineHornung2004LaplacianSE} to formulate an energy term $\mathbb{E}_{def}$. The key idea is to penalize the over-exaggerated area by evaluating their corresponding displacements in normal face models. Denote the residual blendshapes of $\{\mathbf{S}(\mathbf{e_{i}})\}_{i=0}^{46}$ and $\{\tilde{\mathbf{S}}(\mathbf{e_{i}})\}_{i=0}^{46}$ by 
\begin{equation}\label{eq:residual_blendshape}
    \mathbf{D}(\mathbf{e_i})=\mathbf{S}(\mathbf{e_i})-\mathbf{S}(\mathbf{e_{x}}),~~~ i=0,1,..., 46,
\end{equation}
\begin{equation}
    \tilde{\mathbf{D}}(\mathbf{e_i})=\tilde{\mathbf{S}}(\mathbf{e_i})-\tilde{\mathbf{S}}(\mathbf{e_{x}}),~~~ i=0,1,..., 46.
\end{equation}
Then we compute a displacement intensity mask $\mathbf{M}(\mathbf{e_i})=\{\mathbf{m_i^n}\}_{n=0}^N$ by normalizing each vertex displacement vector $\tilde{\mathbf{d}}_i^n\in \tilde{\mathbf{D}}(\mathbf{e_i})$ into a scalar $\mathbf{m}_i^n\in [0,1]$:
\begin{equation}
    \mathbf{m}_i^n=\frac{\|\tilde{\mathbf{d}}_i^n\|}{\max_{1\leq n\leq N}\|\tilde{\mathbf{d}}_i^n\|},~~~ 
    n=0, 1,..., N,
\end{equation}
where $N$ is the number of dense mesh vertices for both $\tilde{\mathbf{S}}(\mathbf{e_i})$ and $\mathbf{S}(\mathbf{e_i})$, $i=0,..,46$. 

Let $\mathbf{D^\star}=\{\mathbf{D^\star}(\mathbf{e_i}\}_{i=0}^{46}$ be the optimizing target. We define $\mathbb{E}_{def}$ as a weighted Laplacian deformation term:
\begin{equation} \label{eq:E_def}
    \mathbb{E}_{def}(\mathbf{D^\star})=\sum_{i=0}^{46}\|\Delta (\mathbf{D^\star}(\mathbf{e_i})-\mathbf{M}(\mathbf{e_i})\cdot \mathbf{D}(\mathbf{e_i}))\|^2.
\end{equation}
In practice, we employ the standard cotangent weights~\cite{Pinkall1993ComputingDM} for the Laplacian operator $\Delta(*)$ defined on discrete mesh vertices. Eq.\ref{eq:E_def} essentially is to use the deformation in normal face domain to guide the deformation in caricature domain so as to penalize undesired over-exaggeration.

For the second objective, our target is to constrain the two parallel blendshapes $\{\mathbf{S}(\mathbf{e_{i}})\}_{i=0}^{46}$ and $\{\tilde{\mathbf{S}}(\mathbf{e_{i}})\}_{i=0}^{46}$ to keep similar group structure. Inspired by~\cite{Ribera2017FacialRW}, we use the cosine distance to evaluate the similarities among normal face blendshapes:
\begin{equation}
    cos(\mathbf{e_i},\mathbf{e_j})=\frac{\tilde{\mathbf{D}}(\mathbf{e_i})\cdot \tilde{\mathbf{D}}(\mathbf{e_j})}{\|\tilde{\mathbf{D}}(\mathbf{e_i})\|\cdot \|\tilde{\mathbf{D}}(\mathbf{e_j})\|},~~~i,j\in \{0,1,...,46\}.
\end{equation}
Then the correlation weights are applied on caricature blendshapes to enforce the same interior similarities, which leads to our second  energy term:
\begin{equation}
    \mathbb{E}_{str}(\mathbf{D^\star})=\sum_{i=0}^{46}\sum_{j=0}^{46}cos(\mathbf{e_i},\mathbf{e_j})\|\mathbf{D^\star}(\mathbf{e_i})-\mathbf{D^\star}(\mathbf{e_j})\|^2.
\end{equation}

Finally, we add energy term $\mathbb{E}_{smo}$ to improve the local smoothness for optimized caricature blendshapes:
\begin{equation}
    \mathbb{E}_{smo}(\mathbf{D^\star})=\sum_{i=0}^{46}\|\Delta (\mathbf{D^\star}(\mathbf{e_i}))\|^2.
\end{equation}

Summing the three terms up gives our full objective function of caricature blendshape optimization:
\begin{equation}\label{eq:optimization}
    \mathbb{E}_{total}=\lambda_{def}\mathbb{E}_{def}+\lambda_{str}\mathbb{E}_{str}+\lambda_{smo}\mathbb{E}_{smo},
\end{equation}
where $\lambda_{def}$, $\lambda_{str}$ and $\lambda_{smo}$ are trade-off weights. The entire optimization can be solved linearly. Once $\mathbf{D^\star}$ is optimized, the optimized caricature blendshapes $\mathbf{S^\star}$ can be obtained by adding the source model back according to Eq.~\ref{eq:residual_blendshape}.

\begin{figure}
\begin{center}
  \includegraphics[width=1\linewidth]{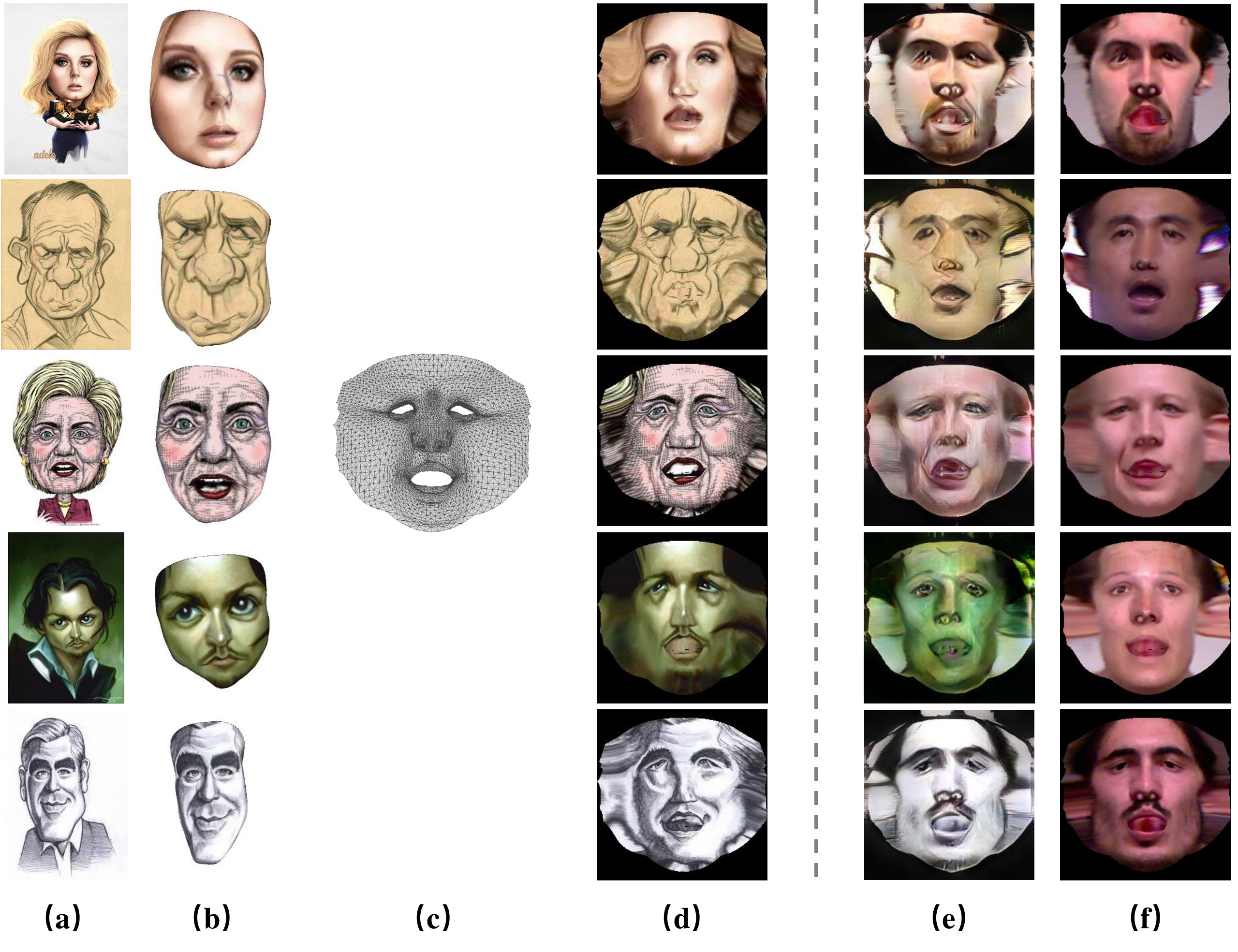}
\end{center}
\vspace{-0.1in}
  \caption{\textbf{Selected texture training data.} (a)-(d) Source caricature image, model, ARAP parameterization and extracted texture; (e)(f) Style-transferred and original human face textures. By transferring image styles, we augment the training dataset with normal face textures.}
\label{fig:texture_sample}
\end{figure}
\vspace{-0.1in}
\subsection{Texture Generation Component}\label{sec:texture_component}
As pointed out in Sec.\ref{sec:overview}, only manipulating the underlying geometric shapes is in general not sufficient to generate realistic caricature expressions. With expression changes, the face images will contain varying transformations in color space, especially in the areas of mouth and eyes. In order to model the specific color transformation, we need to modify the texture map according to the target expression. Let  $\mathbf{T}(\mathbf{e_{x}})\in \mathbb{R}^{H\times W\times 3}$ denote source caricature texture and $\mathbf{e_t}\in \mathbb{R}^D$  be the target expression. The texture model $\mathcal{M}$ seeks to estimate the target texture map as $\mathbf{T}(\mathbf{e_{t}}) = \mathcal{M}(\mathbf{T}(\mathbf{e_{x}}), \mathbf{e_t})$.

Toward this objective, we propose a conditional GAN model. However, training image-based generative networks requires a large-scale dataset, which we do not have for caricatures. To overcome this issue, we augment our caricature texture training data by adapting normal human face textures to various caricature styles. In this way, our network can be trained on both adapted texture dataset (in a supervised manner) and real caricature texture dataset (in an unsupervised manner). 
\vspace{-0.1in}
\subsubsection{Network Structure}\label{sec:network_structure} Fig.~\ref{fig:network_structure} shows the overall network structure of our texture generation model $\mathcal{M}$, which is composed of a two-branch generator $G$ and a discriminator $D$. To condition the texture generation results on different expressions, we randomly pair a source caricature and a target expression as $(\mathbf{T}(\mathbf{e_x}), \mathbf{e_t})$. Following the attention mechanism design in~\cite{Pumarola2018GANimationAF}, the generator $G=(G_A, G_C)$ takes as input $(\mathbf{T}(\mathbf{e_x}), \mathbf{e_t})$ and estimates two masks: color mask $\mathbf{C}\in R^{H\times W\times 3}$ and attention mask $\mathbf{A}\in [0,1]^{H\times W\times 1}$, where the former suggests what color values should be changed to and the latter indicates where to make changes. The final regressed target texture $\mathbf{T}(\mathbf{e_x|e_t})$ is computed by:
\begin{equation}\label{eq:attention}
    \mathbf{T}(\mathbf{e_x|e_t}) = \mathbf{A}\cdot \mathbf{T}(\mathbf{e_x}) + (1-\mathbf{A})\cdot C .
\end{equation}
Different from the \textit{GANimation} model~\cite{Pumarola2018GANimationAF}, which operates in image domain, our texture generation operates on texture map $\mathbf{T}$, which is universally aligned and focuses on texture attributes only. Note that our attention estimation branch is based on source and target expression coefficients and does not need to consider the color information of $\mathbf{T}(\mathbf{e_x})$. 

The discriminator $D$ is responsible for evaluating the expression of the generated texture map and discriminating real and fake textures. In one branch of $D$, a Patch-GAN~\cite{Isola2016ImagetoImageTW} instance will map an input image to overlapping patches and output real probability for each patch. The other branch of $D$ will regress the output texture to estimate its expression. 

\vspace{-0.1in}
\subsubsection{Training}
\noindent \textbf{Data Acquisition}. Our training data comes from two sources. The first one is from real caricature images. We collect thousands of caricature paintings from internet and reconstruct their 3D shapes with manually labeled landmarks. Then by applying rasterization pipeline on a pre-defined ARAP parameterization map, we obtain their individual textures. Some selected examples are shown in Fig.~\ref{fig:texture_sample}. Due to the big data requirement of deep learning methods, we also explore the possibility of augmenting our training data from normal face textures. We use the \textit{Adaptive Instance Normalization}~\cite{Huang2017ArbitraryST} approach to transfer normal face textures in FaceWarehouse~\cite{cao2014facewarehouse} dataset into random caricature style. The augmented data samples are also given in Fig.~\ref{fig:texture_sample}. It can be observed that in this way, the transferred textures look very similar to the original samples.

\noindent \textbf{Supervised Training}. In FaceWarehouse~\cite{cao2014facewarehouse} dataset, there are multiple expressions of the same person. We make use of this feature to train the texture model in a supervised manner with source texture map $\mathbf{T}(\mathbf{e_x})$, target texture map $\mathbf{T}(\mathbf{e_t})$ and `ground-truth' attention map $\mathbf{A}(\mathbf{e_x, e_t})$, which is obtained by mapping the vertex displacement map computed from the corresponding 3D shapes to the texture map space. In other words, a large 3D vertex displacement at a 3D vertex suggests likely texture color change at the corresponding texture map location. 

We train the generator $G$ with two loss functions. The first one is attention mask regression loss. The generated attention $\mathbf{A}(\mathbf{e_x|e_t})$ should be regressed close to the pre-computed map $\mathbf{A}(\mathbf{e_x, e_t})$:
\begin{equation}
    \mathbb{L}_{att}=\|\mathbf{A}(\mathbf{e_x|e_t})-\mathbf{A}(\mathbf{e_x, e_t})\|_1 .
\end{equation}
The second loss is color transformation loss. The generated texture map $\mathbf{T}(\mathbf{e_x|e_t})$ combined by Eq.~\ref{eq:attention} should be close to the groundtruth $\mathbf{T}(\mathbf{e_t})$:
\begin{equation}
    \mathbb{L}_{color}=\|\mathbf{T}(\mathbf{e_x|e_t})-\mathbf{T}(\mathbf{e_t})\|_1 .
\end{equation}

\noindent \textbf{Unsupervised Training}. For real caricature data, there is no paired expressions of the same identity. Thus, we do not have groundtruth for either attention mask or output texture. So we replace the supervised training loss with cycle reconstruction loss: 

\begin{equation}
    \mathbb{L}_{cycle}=\|G(G(\mathbf{T}(\mathbf{e_x})|\mathbf{e_t})|\mathbf{e_x})-\mathbf{T}(\mathbf{e_x})\|_1,
\end{equation}
which means that the generator $G$ is expected to transform source texture map $\mathbf{T(e_x)}$ under target condition $\mathbf{e_t}$ and then transform it back under source condition $\mathbf{e_x}$.

\noindent \textbf{Discriminator}. The generators in both scenarios need to cooperate with the discriminator for adversarial training. $D$ is trained with two discriminative loss functions. 

The first one is image adversarial loss $\mathbb{L}_{adv}$, 
\begin{equation}
    \mathbb{L}_{adv}=\mathbb{E}_{\mathbf{T}_{f}\sim\mathbf{P}_{f}}[D(\mathbf{T}_{f})]-\mathbb{E}_{\mathbf{T}_{r}\sim\mathbf{P}_{r}}[D(\mathbf{T}_{r})]+\lambda_{gp}\mathbb{E}_{GP},
\end{equation}
which aims to maximize the probability of classifying the real sample $\mathbf{T}_{r}\sim\mathbf{P}_{r}$ and the fake generation $\mathbf{T}_{f}\sim\mathbf{P}_{f}$ based on the continuous Earth Mover Distance~\cite{Arjovsky2017WassersteinGA}, and $\lambda_{gp}$ is the weight for gradient penalty term $\mathbb{E}_{GP}$ in~\cite{Arjovsky2017WassersteinGA}. The generator $G$ tries to fool $D$ simultaneously.

The second one is expression condition loss $\mathbb{L}_{exp}$, which forces the network to minimize the error between target expression coefficients and regressed ones:
\begin{equation}
    \mathbb{L}_{exp}=\|D(G(\mathbf{T}(\mathbf{e_x})|\mathbf{e_t})) - \mathbf{e_t}\|^2+\|D(\mathbf{T}(\mathbf{e_x})) - \mathbf{e_x}\|^2 .
\end{equation}
Finally, the total loss function becomes:
\begin{equation}
    \mathbb{L}=\lambda_{att}\mathbb{L}_{att} + \lambda_{color}\mathbb{L}_{color} + \lambda_{cycle}\mathbb{L}_{cycle} + \lambda_{adv}\mathbb{L}_{adv} + \lambda_{exp}\mathbb{L}_{exp}.
\end{equation}
All $\lambda$ are hyper-parameters to balance training weights and switch supervised/unsupervised training mode.

\begin{figure}
\begin{center}
  \includegraphics[width=1\linewidth]{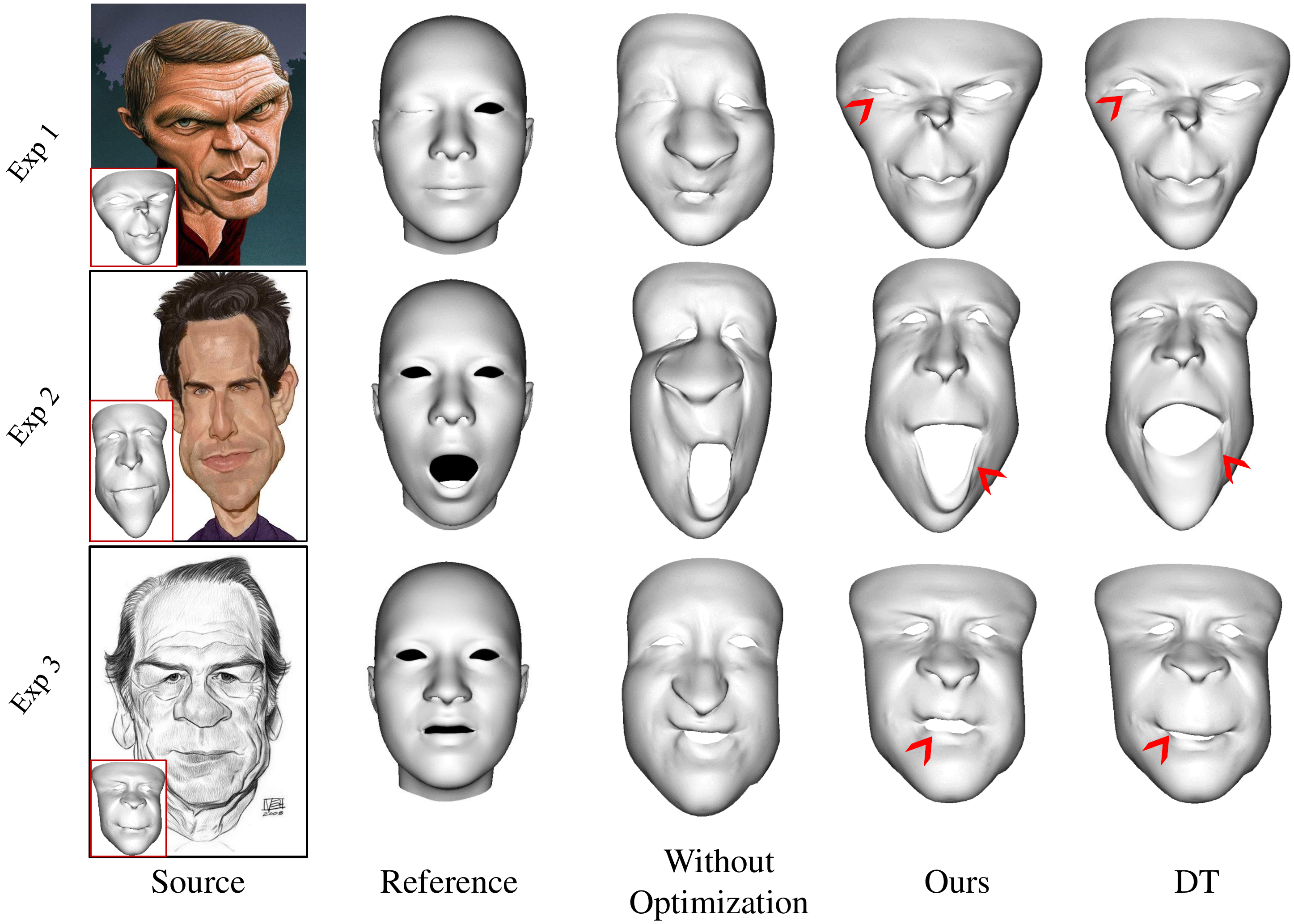}
\end{center}
\vspace{-0.1in}
  \caption{Ablation study on our blendshape optimization method and comparison with \textit{Deformation transfer}~\cite{Sumner2004DeformationTF} (DT). For each source caricature shape in the left column, we show the generated blendshapes of different methods in columns 3-5 with reference to human expressions in the second column. The results clearly indicate that without optimization, the caricature identity will be lost. The red arrows amplify the key different areas between \textit{deformation transfer} and our results. Our method achieves more exaggerated and accurate expression deformations than DT.} 
\label{fig:shape_comparison}
\end{figure}
\begin{figure}
\begin{center}
  \includegraphics[width=1\linewidth]{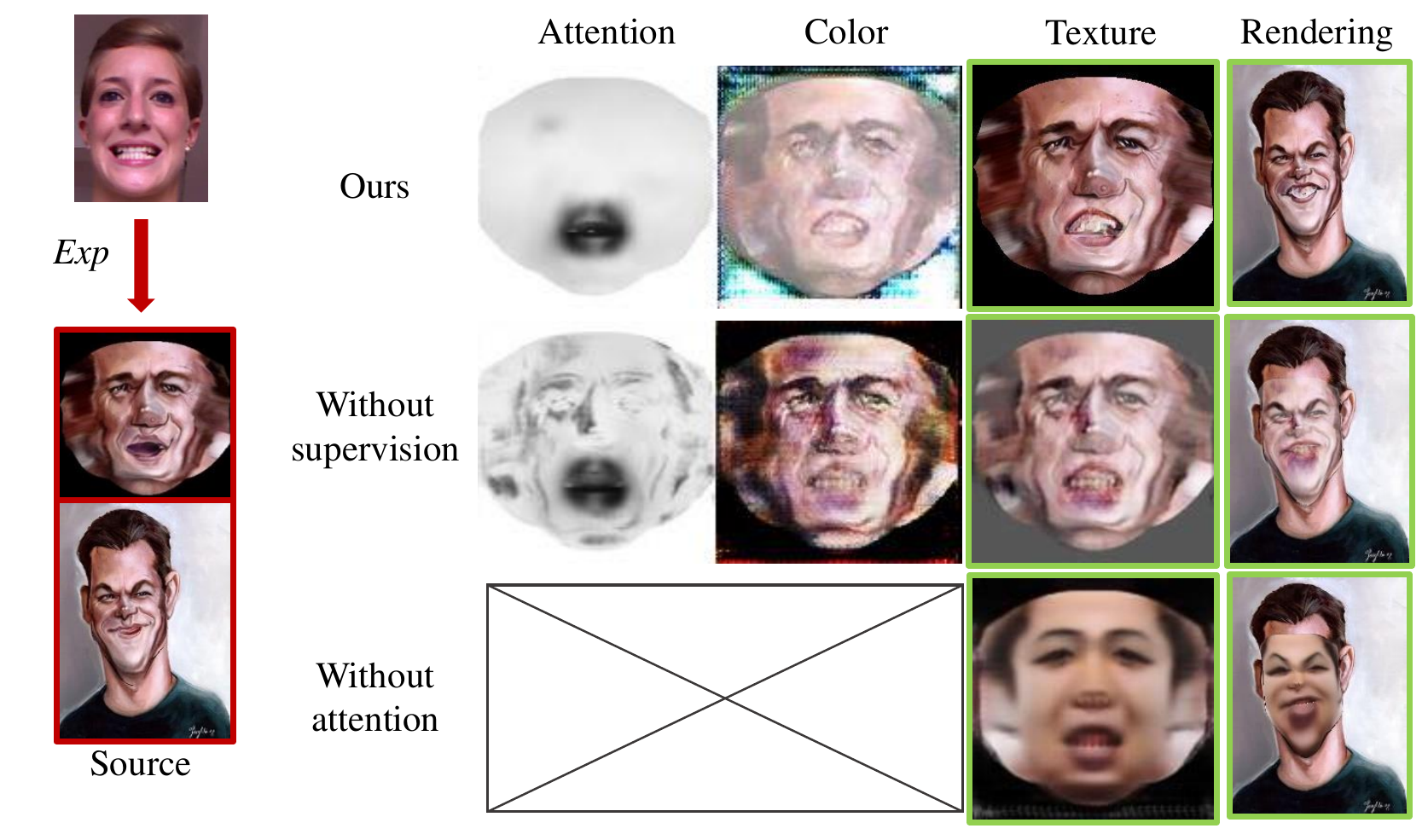}
\end{center}
\vspace{-0.1in}
  \caption{Ablation study on the texture generation component. We compare our method with models trained without fully supervised data or without the attention module. The target expression of mouth-open requires the texture model to generate the teeth part in the mouth area.} 
\label{fig:texture_comparison}
\end{figure}
\begin{figure*}
\begin{center}
  \includegraphics[width=1\linewidth]{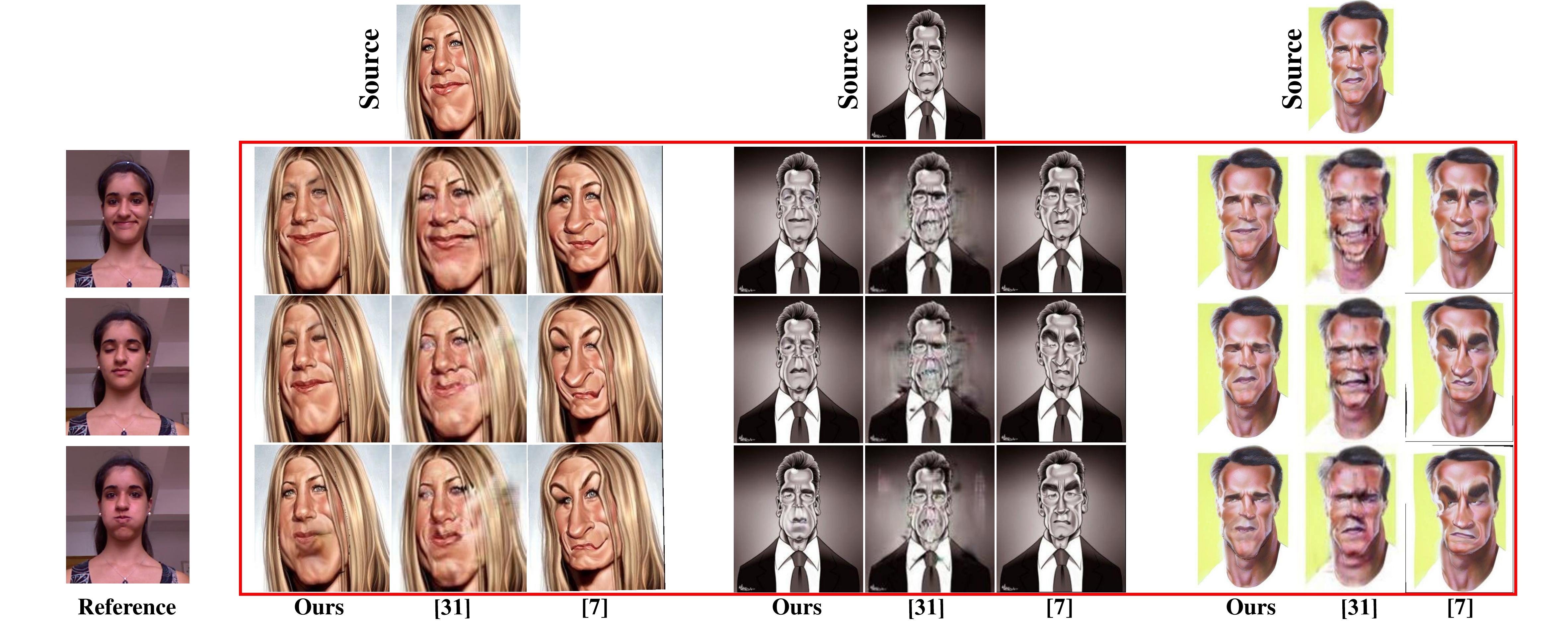}
\end{center}
\vspace{-0.1in}
  \caption{Comparing our method with GANimation~\cite{Pumarola2018GANimationAF} and CariGeoGAN~\cite{Cao2018CariGANsUP}. The reference photos guide the caricature generation by giving 3DMM expression coefficients to~\cite{Cao2018CariGANsUP} and our model or giving Action Unit code to GANimation~\cite{Pumarola2018GANimationAF}. The comparisons show that our results preserve better image quality than~\cite{Pumarola2018GANimationAF} and maintain more stable identity than~\cite{Cao2018CariGANsUP}. Please zoom in for details.}
\label{fig:baseline_comparison}
\end{figure*}
\section{Experiments}\label{sec:experiment}

\subsection{Implementation Details}
The computation of our shape modelling component is conducted on a PC with Intel Xeon W-2133 CPU, 16GB RAM. For caricature blendshape optimization, we set $\lambda_{def}$ as $1.0$, $\lambda_{str}$ as $0.1$ and $\lambda_{smo}$ as 0.05. The full computation of the shape modelling component costs about 1 second. The dynamic texture model is trained on an NVIDIA TITAN V graphics card for about 10 hours. We first train the model in a supervised manner with $\lambda_{att}=10.0, \lambda_{color}=10.0, \lambda_{cycle}=0.0, \lambda_{adv}=1.0$ and $\lambda_{exp}=100.0$. After the attention mask loss converges, we switch to unsupervised training with $\lambda_{att}=0.0, \lambda_{color}=0.0$ and $\lambda_{cycle}=10.0$. In total, we trained the network for 600 epochs with batch size of 40 and learning rate of 1e-5. 
\vspace{-0.1in}
\subsection{Ablation Study}~\label{sec:ablation_study}
\vspace{-0.2in}
\subsubsection{Caricature Blendshapes}
To evaluate the caricature shape model, we compare the blendshapes generated with and without our optimization approach. Besides, we also compare with \textit{deformation transfer}~\cite{Sumner2004DeformationTF}, which uses template model of source and target objects to compute the shape correspondence. Then by directly transferring the deformation of source object, it can generate deformed results for target. In experiments, we make the fitted 3DMM model $\tilde{\mathbf{S}}(\mathbf{e_{x}})$ as source template and caricature model $\mathbf{S}(\mathbf{e_{x}})$ as target template. The other 3DMM expressions $\{\tilde{\mathbf{S}}(\mathbf{e_{i}})\}$ give the deformation reference to drive target caricature shapes. 

Fig.~\ref{fig:shape_comparison} shows some comparison results generated with/without optimization and \textit{deformation transfer}. It can be observed that without optimization, although the generated caricature shapes can maintain reference expressions, they lose source identity information significantly. In contrast, our method preserves the identitiy information well. Compared with \textit{deformation transfer}, our method achieves more exaggerated and accurate expression deformations. This is because \textit{deformation transfer} only copies deformations of normal face shapes and they are not strong enough to animate caricature shapes. In contrast, our method adopts the cross-domain translation so that we can handle the exaggeration by learning with various expressions of different caricatures as domain knowledge.
\begin{figure*}
\begin{center}
  \includegraphics[width=1\linewidth]{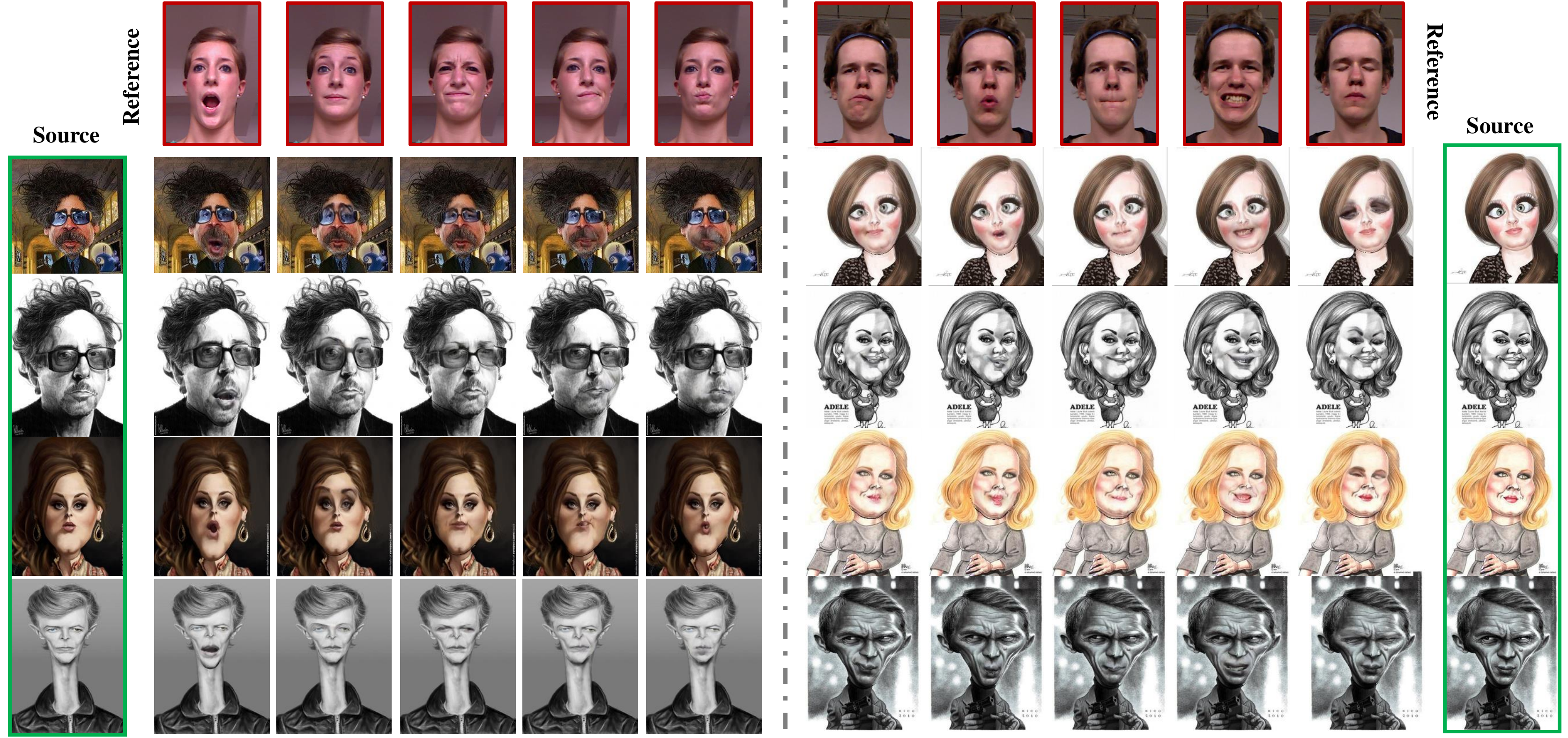}
\end{center}
\vspace{-0.1in}
  \caption{Rendering expression editing examples. Images in first row (red block) are references that specify target expressions. For each caricature image, we only need to generate its optimized blendshapes once and they can be used to produce many caricature images with new expressions by simple blending specified in Eq.~\ref{eq:Se_r}. Our model can handle various expressions such as open-mouth, pouting, contempt and closing-eyes. Please zoom in for details.}
\label{fig:gallery}
\end{figure*}
\vspace{-0.1in}
\subsubsection{Dynamic Texture Generation}
We evaluate our trained conditional texture generation network with two ablation studies. First, we remove the adapted texture dataset from normal face textures, which leaves the entire training to be purely unsupervised. Specifically, we keep the unsupervised losses $\mathbb{L}_{cycle}, \mathbb{L}_{adv}$ and $\mathbb{L}_{exp}$ and train the network only on original caricature texture dataset. The visual comparison in Fig.~\ref{fig:texture_comparison} proves the significance of the adapted dataset and the combining training scheme. Without the supervised training, the conditional GAN model is hard to predict satisfactory color transformation and attention mask.

The second study is on the attention mechanism. We compare our model with and without the attention branch. Note that removing the attention branch makes our model degenerate to a vanilla cGAN network. Fig.~\ref{fig:texture_comparison} shows that the results of our model without the attention is poor.

The main reason is that the generator is hard to handle various image styles and color transformations simultaneously and the attention mechanism can well ease that difficulty by estimating color and attention mask in a separate way.

\subsection{Comparisons}~\label{sec:baseline_comparison}
In this part, we evaluate our full model through comparison with other competitive methods. Since there is no face editing works proposed specifically for caricatures, we choose two feasible methods, GANimation~\cite{Pumarola2018GANimationAF} and CariGeoGAN~\cite{Cao2018CariGANsUP}, as our baselines. 
\vspace{-0.1in}
\subsubsection{Comparison with GANimation~\cite{Pumarola2018GANimationAF}}
As one of the \textit{state-of-the-art} facial expression editing works, GANimation~\cite{Pumarola2018GANimationAF} is capable of generating plausible human portraits conditioned on facial Action Unit~\cite{Ekman1978FacialAC}, which is similar to our problem settings. So we compare with GANimation~\cite{Pumarola2018GANimationAF} to see if it can be generalized to caricature images. We directly use the pre-trained model of~\cite{Pumarola2018GANimationAF} and make the GANimation and our model target for the same expression reference. Fig.~\ref{fig:baseline_comparison} shows some samples generated by both methods. It can be clearly seen that our results are of much better image quality. The reason is that we process caricature images more 
meticulously with disentangled 3D geometry and texture components, while GANimation is built entirely on 2D domain. 
\subsubsection{Comparison with CariGeoGAN~\cite{Cao2018CariGANsUP}}
CariGeoGAN~\cite{Cao2018CariGANsUP} is cycle consistent mapping which translates landmarks between normal face and caricature. As aforementioned, the caricature landmarks can be translated to normal face landmarks by~\cite{Cao2018CariGANsUP}, which can be used to fit 3DMM model with expression parameters. The 3DMM-based blendshapes are then translated back to caricature domain to form a group of caricature landmarks, which are capable of composing arbitrary new expressions given 3DMM parameters. Consequently, we use 2D thin-plate spline warping~\cite{Bookstein1989PrincipalWT} to deform source caricature images guided by target landmarks. The examples illustrated in Fig.~\ref{fig:baseline_comparison} show that this baseline could not preserve the identity information compared with our method. The underlying explanation is that our shape model is optimized on dense mesh vertices but CariGeoGAN~\cite{Cao2018CariGANsUP} only warps the image with sparse landmarks, which likely results in losing identity information in such low-dimensional space. 
\vspace{-0.1in}
\subsection{More Results}~\label{sec:more_results}
Our caricature expression model can be used to edit the expressions of any given caricature painting. Since the shape and texture components are conditioned upon the same facial expression representation (3DMM model), our method can be easily integrated with previous 3DMM-based human face reconstruction works by extracting their expression coefficients and approximating blendshape weights.

For a caricature, once its blendshapes $\{\mathbf{S}(\mathbf{e_{i}})\}_{i=0}^{46}$ are optimized, it can be used to easily compute another shape under random input expression parameters ${\mathbf{e_r}}$ by regressing blendshape weights ${\mathbf{w_r}}\in \mathbb{R}^{46}$ on corresponding normal face blendshapes $\{\tilde{\mathbf{S}}(\mathbf{e_{i}})\}_{i=0}^{46}$ and mapping ${\mathbf{w_r}}$ to caricature as:

\begin{equation} \label{eq:Se_r}
        \mathop{\arg\min_{\mathbf{w_r}}~~~\|\tilde{\mathbf{S}}(\mathbf{e_{r}})-\tilde{\mathbf{S}}(\mathbf{e_{0}})-\sum_{i=1}^{46}\mathbf{w_r^i}(\tilde{\mathbf{S}}(\mathbf{e_{i}})-\tilde{\mathbf{S}}(\mathbf{e_{0}}))\|^2}
\end{equation}
\begin{equation}
     \mathbf{S(e_r)}=\mathbf{S(e_0)}+\sum_{i=1}^{46}\mathbf{w_r^i}(\mathbf{S(e_i)}-\mathbf{S(e_0)})
\end{equation}
The texture map can be quickly inferred by applying the trained texture generation network as $\mathbf{T(e_r)}=G(\mathbf{T(e_x)}|\mathbf{e_r})$.
Finally, we reuse the camera projection matrix approximated in Sec.~\ref{sec:preliminary} to render new caricature images. Fig.~\ref{fig:gallery} shows a gallery of expression editing results generated from real caricature paintings. More results can be found in the supplementary material.

\section{Conclusion}\label{sec:conclusion}
We have presented a method to model caricature expression, which extends traditional 3DMM representation to caricature domain. The method has shape and texture components, for which geometric optimization and deep learning methods are developed, respectively. By the method, a group of 3D blendshapes and a dynamic texture generator are cooperated so that one can transform a caricature image into arbitrary expressions. Hence one can easily manipulate the expressions of artist-drawn caricature painting while preserving the exaggerated identity. 
In future, more applications such as video-driven caricature animation will be developed with sequential consistency constraints.

\begin{acks}
This research is partially supported by the National Natural Science Foundation of China (No. 61672481), Youth Innovation Promotion Association CAS (No. 2018495), Zhejiang Lab (NO. 2019NB0AB03), NTU DSAIR grant (No. 04INS000518C130), the National Research Foundation, Singapore under its International Research Centres in Singapore Funding Initiative, and Monash FIT Start-up Grant.
\end{acks}

\bibliographystyle{ACM-Reference-Format}
\bibliography{egbib}

\end{document}